\documentclass{article}
\usepackage{amsmath,graphicx,mlspconf}

\usepackage{algorithm}
\usepackage{algorithmic}

\usepackage{hyperref}       
\usepackage{url}            
\usepackage{amsfonts}       

\usepackage{mathtools} 

\hypersetup{breaklinks=true,
            colorlinks=true,
            linkcolor=black,
            citecolor=black,
            filecolor=black,
            urlcolor=black}

\usepackage{microtype}

\usepackage{tikz} 
\usepackage{tikzscale}

\usepackage{bbm}

\newcommand{\given} { \,|\, }

\newcommand{\tr}{^{\text{T}} }

\newcommand*\diff{\mathop{}\!\mathrm{d}}

\newcommand{\x}{\mathbf{x}}
\newcommand{\f}{\mathbf{f}}
\newcommand{\y}{\mathbf{y}}
\newcommand{\X}{\mathbf{X}}
\newcommand{\F}{\mathbf{F}}
\newcommand{\Y}{\mathbf{Y}}
\newcommand{\C}{\mathbf{C}}
\newcommand{\id}{\text{d}}
\newcommand{\Normal}[1] { \mathrm{N} {\left(#1\right)}  }

\DeclareMathOperator*{\argmin}{arg\,min}

\DeclareMathOperator{\E}{\mathbb{E}}

\usepackage{xspace}
\newcommand{\acr}[1]{\uppercase{#1}\xspace}
\newcommand{\bo}{\acr{bo}}
\newcommand{\ep}{\acr{ep}}
\newcommand{\qei}{{\small q}-\acr{ei}}

\newcommand{\ts}{\acr{ts}}

\newcommand{\mfvi}{\acr{vi}}
\newcommand{\mcmc}{\acr{mcmc}}
\newcommand{\vi}{\acr{vi}}
\newcommand{\gp}{\acr{gp}}
\newcommand{\ei}{\acr{ei}}

\newcommand{\gps}{\textsc{gp}s}
\newcommand{\pbbo}{\acr{pbbo}}

\newcommand{\kl}{\acr{kl}}

\newcommand{\hmc}{\acr{hmc}}

\usepackage{pgfplotstable}

\title{PREFERENTIAL BATCH BAYESIAN OPTIMIZATION}

\name{%
\begin{tabular}{c}
    Eero Siivola$^{\star}$%
    \qquad Akash Kumar Dhaka$^{\star}$%
    \qquad Michael Riis Andersen$^{\dagger}$%
    \qquad Javier Gonz\'{a}lez$^{\ddag}$\thanks{Work done prior to Javier Gonz\'{a}lez joining Microsoft Research.} \\
    \qquad Pablo Garc\'{i}a Moreno$^{\mathsection}$%
    \qquad Aki Vehtari$^{\star}$%
\end{tabular}
}
\address{%
    $^{\star}$ Aalto University $^{\dagger}$ Technical University of Denmark %
    $^{\ddag}$ Microsoft Research $^{\mathsection}$ Amazon.com \\%
}

\expandafter\def\expandafter\UrlBreaks\expandafter{\UrlBreaks
  \do\a\do\b\do\c\do\d\do\e\do\f\do\g\do\h\do\i\do\j%
  \do\k\do\l\do\m\do\n\do\o\do\p\do\q\do\r\do\s\do\t%
  \do\u\do\v\do\w\do\x\do\y\do\z\do\A\do\B\do\C\do\D%
  \do\E\do\F\do\G\do\H\do\I\do\J\do\K\do\L\do\M\do\N%
  \do\O\do\P\do\Q\do\R\do\S\do\T\do\U\do\V\do\W\do\X%
  \do\Y\do\Z}

\begin{document}

\maketitle

\begin{abstract}
Most research in Bayesian optimization (\bo) has focused on \emph{direct feedback} scenarios, where one has access to exact values of some expensive-to-evaluate objective. This direction has been mainly driven by the use of \bo in machine learning hyper-parameter configuration problems. However, in domains such as modelling human preferences, A/B tests, or recommender systems, there is a need for methods that can replace direct feedback with \emph{preferential feedback}, obtained via rankings or pairwise comparisons. In this work, we present preferential batch Bayesian optimization (\pbbo), a new framework that allows finding the optimum of a latent function of interest, given any type of parallel preferential feedback for a group of two or more points. We do so by using a Gaussian process model with a likelihood specially designed to enable parallel and efficient data collection mechanisms, which are key in modern machine learning. We show how the acquisitions developed under this framework generalize and augment previous approaches in Bayesian optimization, expanding the use of these techniques to a wider range of domains. An extensive simulation study shows the benefits of this approach, both with simulated functions and four real data sets.
\end{abstract}
\begin{keywords}
Gaussian processes, Bayesian optimization 
\end{keywords}
\section{INTRODUCTION}

Understanding and emulating the way intelligent agents make decisions is at the core of what machine learning and artificial intelligent aim to achieve. To fulfil this goal, behavioural features can be learned from demonstrations like   when a robot arm is trained using human-generated examples \cite{NIPS2016_6391}. In many cases, however, the optimality of the instances is questionable. Reinforcement learning, via the explicit definition of some reward, is another approach \cite{Sutton:1998:IRL:551283}. That can be, however, subject to biases. Imagine asking a user of a streaming service to score a movie between zero and ten. Implicitly, this question assumes that she/he has a  sense of the scale in which the new movie is  evaluated, which implies that an exploration of the \emph{movie space} has been already carried out. 

An alternative way to understanding an agent's decisions is to do it via preferences. In the movie example, any two movies can be compared without scale. Also, the best of ten movies can be selected or a group of movies can be ranked from the worst to the best. This feedback, which can be provided without a sense of scale, provides information about the user preferences. Indeed, in prospect theory, studies have demonstrated that humans are better at evaluating differences rather than absolute magnitudes \cite{Kahneman79}. 

In the Bayesian optimization literature (\bo), these ideas have also been studied  in cases where the goal is to learn the optimum of some latent preference function defined in some Euclidean space \cite{preferential2017gonzalez}. Available methods use pairwise comparisons to recover a latent preference function, which in turn is used to make decisions about new queries. Despite the \emph{batch} setting being a natural scenario here, where more than two points in the space are compared simultaneously, it has not yet been carefully studied in the literature. One relevant example for batch feedback is product design, especially in the food industry, where one can only produce a relatively small batch of different products at a time. The quality of products, especially for foods, is usually highly dependent on the time since production. The whole batch is usually best to be evaluated at once and the next batch of products should be designed based on the feedback so far. In this work, we show that mechanisms to propose preferential batches \emph{sequentially} are very useful in practice, but far from trivial to define. 

\subsection{Problem formulation}
Let $f: \mathcal{ X} \to \mathbb{R}$ be a well-behaved \emph{black-box} function defined on a bounded subset $\mathcal{X} \subseteq \mathbb{R}^{d}$. We are interested in solving the global optimization problem of finding 
\begin{equation}\label{eq:problem}
\x_{\text{min}} = \argmin_{\x \in \mathcal{X}} f(\x).
\end{equation}

We assume that $f$ is not directly accessible and that (noisy) queries to $f$ can only be done in batches $\mathcal{B} = \{\x_i \in \mathcal{X} \}_{i=1}^{q}$. Let $f$ be evaluated at all the batch locations, $y(\x_i) = f(\x_i) + \epsilon$, $i=1,\ldots,q$, where $\epsilon \sim \Normal{0, \sigma^2}$ is a random noise with variance $\sigma^2$. We assume that we can receive a set of pairwise preferences of the noisy evaluations on the batch. Here a pairwise preference is defined by $\x_i \prec \x_j := y(\x_i) \leq y(\x_j)$. The goal is to find $\x_{min}$ by limiting the total number of batch queries to $f$, which are assumed to be expensive. This setup is different from the one typically used in \bo where direct feedback from (noisy) evaluations of $f$ is available \cite{Snoek*Larochelle*Adams_2012}.

In particular, we are interested in cases in which the preferential feedback is collected in a sequence of $B$ batches $\mathcal{B}_b = \{\x_i \in \mathcal{X} \}_{i=1}^{q}$ for $b=1,\dots,B$. Within each batch, at iteration $b$, the feedback is assumed to be collected as a complete (or partial) ordering of the elements of the batch, $I \in \mathbb{N}^q$, s.t.  $\x^b_{I_i} \prec \x^b_{I_j},\;\forall i<j\leq q$ or by the selection of the preferred element $\x^b_i$ of batch $ \x^b_i \prec \x^b_j\; \forall j \neq i$.

We concentrate mainly on the batch winner case in this work. We argue that the batch winner feedback is the most useful type of batch feedback. As the preferential feedback is collected from humans, the full ranking of the batch is laborious for large batches and it sometimes even is impossible (e.g. in A/B testing). The full ranking can also be reduced to the batch winner case. Besides, as  we later demonstrate in the experiments, the added benefit of full ranking instead of the batch winner is smaller than the difference between different acquisition functions.

\subsection{Related work and contributions}
Pairwise comparisons are usually called \emph{duels} in the \bo and bandits literature. In the \bo context, \cite{chu2005preference} introduced a likelihood for including preferential feedback into Gaussian processes (\gps). \cite{chu2005extensions} recomputed the model for all possible duel outputs of a discrete dataset and used the expected entropy loss to select the next query. \cite{brochu2010interactive} used expected improvement (\ei) \cite{movckus1975bayesian} sequentially to select the next duel. Most recently, \cite{preferential2017gonzalez} introduced a new state of the art and non-heuristic method inspired by Thompson sampling to select the next duel. 

In this work, we introduce a method called \emph{preferential batch Bayesian optimization} (\pbbo) that allows optimizing black-box functions with \bo when one can query preferences in a batch of input locations. The main contributions are:
\begin{itemize}
\item We formulate the problem so that the model for latent inputs in preference feedback scales beyond a batch size of two.
\item We present and compare two alternative inference methods for the intractable posterior that results from the proposed batch setting.
\item We adapt two well known acquisition functions to the proposed setting.
\item We compare all inference methods, acquisition functions, and batch sizes jointly in extensive experiments with simulated and real data.
\end{itemize}
The code for reproducing the results is available at \url{https://github.com/EmuKit/emukit/tree/master/emukit/examples/preferential_batch_bayesian_optimization}.

The remainder of the paper is organized as follows. In Section~\ref{sec:theory}, we introduce the theoretical background. In Section~\ref{sec:sequential}, we introduce batch input preferential Bayesian optimization and three acquisition functions. In Section~\ref{sec:experiments} we show the benefits of our approach with simulated and real data. We conclude the paper in Section~\ref{sec:conclusion}.

\section{MODELING BATCH PREFERENTIAL FEEDBACK WITH GAUSSIAN PROCESSES} \label{sec:gp}
\label{sec:theory}

We assume that the latent black-box function $f$ is a realization of a zero-mean Gaussian process (\gp), $p(f) = \mathcal{GP}$ fully specified by some covariance function $K$ \cite{rasmussen2006gaussian}. The covariance function specifies the covariance of the latent function between any two points.

\subsection{Likelihood for batches of preferences}
We propose a new likelihood function to capture the comparisons that are collected in batches. Assuming the general case of batch $\mathcal{B}$ of $q$ locations and $m$ preferences in a list $\C \in \mathbb{N}^{m \times 2}$, such that $\x_{C_{i,1}} \prec \x_{C_{i,2}} \forall i \in [1,\ldots,m]$, the likelihood of the preferences is
\begin{align}
 p(\C | \f) =  \int \ldots \int &  \left( \prod_{i=1}^m \mathbbm{1}_{y_{C_{i,1}}  \leq  y_{C_{i,2}}} \right) \label{eq:lik}  \\ &  \left( \prod_{k=1}^q N(y_k| f_k, \sigma^2) \right)  \diff y_1  \ldots \diff y_q, \nonumber
\end{align}
where $f_k$ is latent function value at $\x_k$ and $\sigma$ is the noise of the comparison.
This likelihood takes jointly into account the uncertainty of the preferences. As a special case, for batch size of two, the likelihood reduces to the one introduced in \cite{chu2005preference}.
If the provided feedback is only the batch winner $\x_j$, the likelihood can be further simplified to (see details on the supplementary material)
\begin{equation}
p(\C | \f) =  \int N(y_j| f_j, \sigma^2) \prod_{i=1, i \neq j}^q \Phi \left( \frac{y_j-f_i}{\sigma} \right)  \diff y_j, \label{eq:b-winner}
\end{equation}
where $\Phi(z) = \int_{-\infty}^z \Normal{\gamma | 0,1} \diff \gamma$. As the likelihood is not Gaussian, the posterior distribution is intractable and some posterior approximation has to be used.

\subsection{Posterior distributions}

The posterior distribution and the posterior predictive distributions of the model outcome are needed for making reasoned decisions based on the existing data. Let us assume $B$ preference outcome observations $\C^b\in \mathbb{N}^{m \times 2}$ at batches $\X^b \in \mathbb{R}^{q \times d}$ ($b=1,\ldots,B$). Let us assume that the unknown latent function values $\f^b \in \mathbb{R}^{q \times 1}$ ($b=1,\ldots,B$) have a \gp prior and each batch of preferences is conditionally independent given the latent values $\f^b$ at $\X^b$.
The joint posterior distribution of all the latent function values $\{\f^p_b\}_{b=1}^B$ and $\f^*$ (at unseen locations $\X^*$) is
\begin{align}
 p( \f^*, \{\f^b \}_{b=1}^B | & \X^*, \{\X^b \}_{b=1}^B, \{\C^b\}_{b=1}^B) \propto \label{eq:posterior}\\ &  p(\f^*, \{\f^b \}_{b=1}^B  | \X^*, \{\X^b  \}_{b=1}^B) \prod_{b=1} ^B p(\C^b | \f^b )  . \nonumber 
\end{align}
The posterior predictive distribution for $\f^*$ is obtained by integrating over $\{ \f^b \}_{b=1}^B$. 

\subsection{Model selection and inference}
Since the likelihood of the preferential observations is not Gaussian, the whole posterior distribution is intractable and some approximation has to be used. Next, we present expectation propagation (\ep) and variational inference (\vi) approximations. \ep can be used for general batch feedback in Eq.~\eqref{eq:lik}. With \vi we limit to the batch winner case in Eq.~\eqref{eq:b-winner}. See more details on both these methods in the supplementary material.

\subsubsection{Expectation propagation using multivariate normal as an approximate distribution \label{sec:epmodel}}
\ep \cite{minka2001expectation} approximates some intractable likelihood by a distribution from the exponential family so that the Kullback–Leibler (\kl) divergence from the posterior marginals to the approximative posterior marginals is minimized. In this paper, we use multivariate normal distributions for each batch so that in the posterior distribution in Eq.~\eqref{eq:posterior},  $\prod_{b=1}^B p(\C^b | \f^b )$ is approximated by $\prod_{b=1} ^B N(\f ^b | \boldsymbol{\mu}^b, \mathbf{\Sigma}^b )$. In practice, for approximative distributions from the exponential family, this can be done in an iterative manner where the approximation of batch $b$ is replaced by the original one and the approximation  of batch $b$ is updated by matching the moments the full approximative distribution and the replaced one. Since the moments for the distribution in Eq.~$\eqref{eq:lik}$ are not analytically available, we approximate them by sampling.

\subsubsection{Variational Inference using stochastic gradient descent \label{sec:vimodel}}
The batch winner likelihood (Eq.~\eqref{eq:b-winner}) has the same structure as the one-vs-each likelihood in the context of multiclass classification with linear models \cite{titsias16}. In this formulation, the product of pairwise comparisons is used as a lower bound for the log of the batch winner likelihood,
\begin{eqnarray}
&\log \int N(y_j| f_j, \sigma^2) \prod_{i=1, i \neq j}^q \Phi \left( \frac{y_j-f_i}{\sigma} \right)  \diff y_j \nonumber\\ &\geq \sum_{i=1, i \neq j}^q \log \Phi\left(\frac{f_j - f_i}{\sqrt{\sigma_j +\sigma_i}}\right).
\end{eqnarray}
We want to highlight that this is not an exact likelihood, but a lower bound since we do not integrate over the uncertainty of the batch winner and we thus ignore the dependency of the observations within a batch.

Let $\mathbf{K}$ be the prior covariance matrix at $\X = [\X^1,\ldots,\X^B]\tr$, let  $\boldsymbol{\alpha}$ be a vector, and let $\boldsymbol{\beta}$ be another vector. Following~\cite{opper2009variational}, we posit a Gaussian approximation of the posterior,
\begin{equation}
 q( \f) = {N}(\f|\mathbf{K} \boldsymbol{\alpha},\left( \mathbf{K} +  \mathbf{I} \boldsymbol{\beta} \right)^{-1} ).
 \label{eq:viposterior2}
\end{equation}
The variational parameters are optimised in an inner loop with stochastic gradient descent after collecting derivatives and likelihood terms from the comparison. The benefit of this form compared to \ep is that it gives us a single bound making the optimization easier.

\section{SEQUENTIAL LEARNING FOR BATCH SETTINGS}  \label{sec:sequential}
In this section, we present two strategies for selecting the batch locations. 
Although this work mainly concentrates on the batch winner case, the presented acquisition functions are applicable for the general preferential feedback of Eq.~\eqref{eq:lik}.

\subsection{Expected Improvement for preferential batches} \label{sec:QEI}
Expected improvement is a well established exploitative acquisition function that computes the expected improvement over the minimum of the values observed so far, $y_{min}$. It also has an extension in the batch setting, batch \ei (\qei) \cite{chevalier2013fast}. In the context of preferential feedback, we do not observe the exact function values and do not know the minimum of the observed values. This adds one more source of uncertainty to the \qei for batches of direct feedback (see details on the supplementary material).
One way of avoiding the computational cost of having to integrate over the uncertainty of the minimum and not having to update the model posterior is to use the minimum of the mean of the latent posterior ($\mu_{min} = \min_{ i} \mu (\x_i)$) of the training data as a proxy for $y_{min}$. In this case, the acquisition function equals the relatively fast \qei \cite{chevalier2013fast}
\begin{align}
\text{q-EI}  =& \E_{\y} \left[ \left( \max _{i \in [1, \ldots , q]} \left(\mu_{min}\!\!- y_i \right) \right)_+ \right] \nonumber \\= &  \sum_{i=1}^q  \E_\y \left( \mu_{min} - y_i \, \big |  \, y_i \leq \mu_{min},\,  y_i \leq  y_j \, \forall j \neq i  \right) \nonumber \\ &    p(y_i \leq \mu_{min},\, y_i \leq y_j \, \forall j \neq i ). \label{eq:qei}
\end{align}

\subsection{Thompson sampling for batches} \label{sec:TS}
A purely exploratory approach does not exploit the information about the known good solutions. \ei approaches are known to over-exploit and the proposed approach is very expensive to compute. Although Thompson sampling is heuristic, it is known to work well in practice and nicely balance between exploration and exploitation.
We use the scalable batch \bo approach of \cite{hernandez2017parallel} to select the batch locations in our experiments. In practice, we sample $q$ continuous draws from the posterior predictive distribution of the latent variable and select each batch location as a minimum of the corresponding sample. 

\subsection{Preferential batch Bayesian optimization}
\label{sec:pbbo}
\begin{algorithm}[tb]
\caption{Pseudo-code of the proposed \pbbo method. The inputs are the batch size $q$, the \emph{stopping criterion}, the \emph{acquisition strategy} and the \gp model.
\label{alg:pbbo}}
\begin{algorithmic}[1]
   \WHILE{\emph{stopping criterion} is False}
   \STATE Fit a \gp to the available preferential observations $\{\C^i\}_{i=1}^N$ at $\{\X^i\}_{i=1}^N$.
   \STATE Find $q$ locations $\X^{N+1} = \{\x_i\}_{i=1}^q$ using the \emph{acquisition strategy}.
   \STATE Query the preference $\C^{N+1}$ of $\X^{N+1}$.
   \STATE Augment $\{\X^i\}_{i=1}^N$ with $\X^{N+1}$ and $\{\C^i\}_{i=1}^N$ with $\C^{N+1}$.
   \ENDWHILE
\end{algorithmic}
\end{algorithm}

\begin{figure}[tb!]
	\centering
    \includegraphics{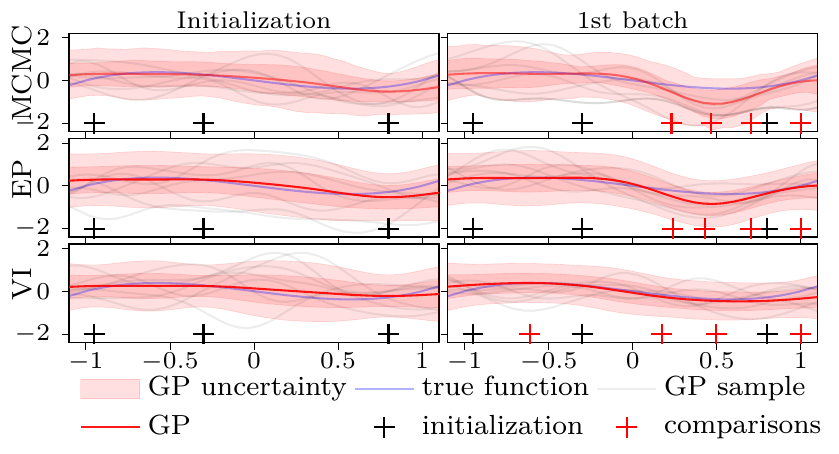}
    \caption{Different rows visualize the true objective and the \gp posterior for different inference methods (Markov chain Monte Carlo (\mcmc), Expectation propagation (\ep), and variational inference (\mfvi)). The first column shows the \gp posterior after observing preferential feedback for a batch of size three (x locations at black '+'-signs). The second column shows the first \bo iteration (x locations at red '+'-signs) using \qei and a batch size of four. The \gp uncertainty is visualized as $\pm$1 and $\pm$2 standard deviations.}
\label{fig:uncertainty}
\end{figure}

The pseudo-code for a general acquisition strategy is presented in Algorithm \ref{alg:pbbo}. The different parts of the algorithm scale as follows. Assuming $N$ batches of size $q$, fitting the \gp has the time complexity of $\mathcal{O} ((Nq)^3)$. The inference method brings some overhead to this.

\section{EXPERIMENTS}
\label{sec:experiments}
From hereafter, \ep stands for the expectation propagation model and \mfvi is the variational inference model. Markov chain Monte Carlo (\mcmc) is used as a ground truth. As a baseline method, we show results if all acquisitions were selected completely at random. The acquisition strategies are abbreviated as follows. \qei stands for the q-expected improvement, and \ts stands for Thompson sampling. The methods are implemented using GPy \cite{gpy2014} and the \mcmc inference is implemented using Stan 2.18.0 \cite{standev2018stan}. In all experiments, the \gp kernel is the squared exponential and the hyper-parameters are fixed to point values by optimizing a regular \gp with 2500 noise free observations. The used \mcmc algorithm is Hamiltonian Monte Carlo (\hmc) with 9000 samples in total from 6 chains. The exact details and more results for all experiments are available in the supplementary material.

\subsection{Effect of the inference method}\label{sec:experiments1}

The different inference methods approximate the uncertainty differently and this affects how the \bo selects the next batch.
Figure \ref{fig:uncertainty} visualizes the \gp posterior for different inference methods with the same training data and then the posterior approximation after the first iteration of \bo when using \qei as the acquisition function. The black-box function is $f(x) = x^3-x.$

The Figure shows that \ep results in very similar posteriors as the \mcmc ground truth when observing only one batch of preferences that are relatively far away from each other (first column). When observing the second batch through \bo, \ep produces a wider posterior than \mcmc, and \mfvi produces a narrower posterior.

\begin{figure}[tb!]
    \centering
    \includegraphics{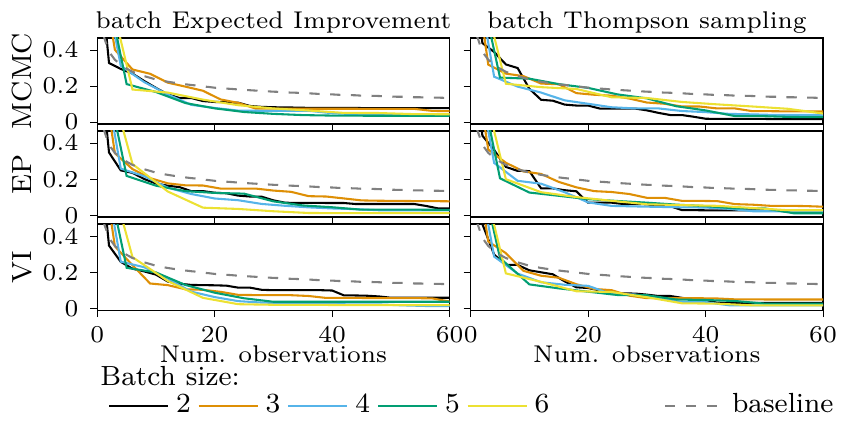}
    
    \caption{Ursem Waves-function from the Sigopt library. Each line illustrates the smallest value of the objective function in the input locations visited so far as a function of the number of observations. The function is scaled between 0 and 1. Different colors in each plot are different batch sizes, rows are different inference methods and columns are different acquisition functions. Each line is a mean of 10 different runs. The dashed black line shows the average performance of the baseline, random search. 
}
\label{fig:Ursem}
\end{figure}

\subsection{Synthetic functions from the Sigopt library} \label{sec:experiments2}

\begin{figure}[tb!]
	\centering
    \includegraphics{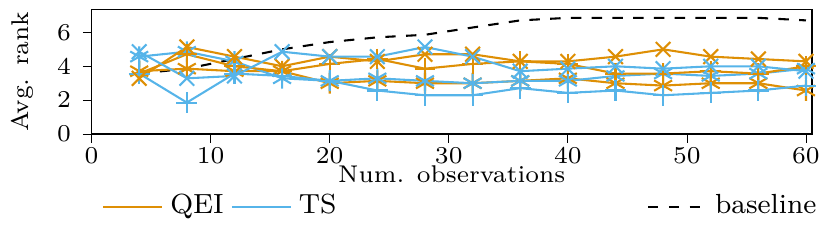} 
    \noindent\\
    \vspace{-2.8mm}
    \hbox{\hspace{10mm}\includegraphics{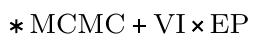}}
\caption{ All combinations of the inference methods and acquisition functions are ranked based on their average performance over 10 runs for the batch size of 4 as a function of the number of evaluations so far. Each combination is given a rank between 1--10 (lower is better) for each iteration. The figure shows the ranks averaged over 6 functions from the Sigopt library (Ursem Waves, Adjiman, Deceptive, MixtureOfGaussians02 and 3 and 4 dimensional Hartmann-functions). Performance of random search is shown as a baseline.}
\label{fig:simulated}
\end{figure}

Sigopt 
library is a collection of benchmark functions developed to evaluate \bo algorithms \cite{dewancker2016stratified}. Ursem Waves is a function from the library with multiple local minima around the search domain. Figure \ref{fig:Ursem} shows the best absolute function value for the locations the function has been evaluated so far as a function of the number of function evaluations. The results are shown for batch sizes 2--6, three acquisition functions, and 3 inference methods. The shown lines are averaged over 10 random runs. The function evaluations are transformed to batch feedback by evaluating the function for the whole batch at once and returning the minimum as the batch winner.
The results show that the both introduced acquisition functions perform better than the baseline. The results show no clear difference between batch sizes for any inference method or acquisition function.
Figure \ref{fig:simulated} shows the average performance of the inference methods and all acquisition functions for the batch size of four. The results are averaged over six functions from the Sigopt library. The results show no clear difference in the performance between \ts and \qei.

\subsection{Comparison of batch winner and full ranking}
\begin{figure}[tb!]
	\centering
    \includegraphics{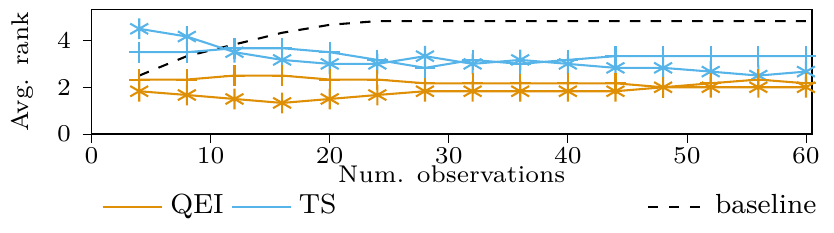}
    \noindent\\
    \vspace{-2.8mm}
    \hbox{\hspace{10mm}\includegraphics{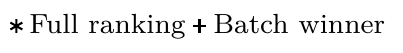}}
\caption{Same as in Figure \ref{fig:simulated}, but for comparing full ranking and batch winner feedbacks.}
\label{fig:comparison}
\end{figure}

One argument against using the batch winner type feedback in \bo is that it is less informative than full ranking. Although it is not always possible to get the full ranking and providing the batch winner is less work, it is interesting to compare these feedbacks. Figure \ref{fig:comparison} shows the average performance of complete ordering and batch winner type feedbacks for both acquisition functions with batch size four. The results are averaged over six functions from the Sigopt library. The results show that the variation in the performances between the acquisition functions is larger than between the feedback types.

\subsection{Real life data case studies} \label{sec:experiments3}

To get insight into how the presented \bo approach performs in real-life applications, we compare the methods also with real data. As the emphasis of this paper is on the batch size, we stick to low dimensional datasets ($d \leq 4$) and present results for batch sizes up to 6. We want to highlight that larger batch sizes and dimensions would remain feasible for \ts. The presented datasets are as follows. Sushi dataset\footnote{Available at: \url{http://kamishima.net/sushi/}} has a complete ranking of 100 sushi items by humans and 4 continuous features describing each sushi item. In the Candy dataset, an online survey was used to collect pairwise preferences to 86 different candies\footnote{Available at: \url{https://github.com/fivethirtyeight/data/tree/master/candy-power-ranking}} with two continuous features. 
We first recovered the full ranking (from best to worst) of the whole dataset and used that to provide feedback to any batch. Since real data is discrete, we use linear extrapolation to compute ranking for points that are not in the dataset.

\begin{figure}[tb!]
    \centering
    \includegraphics{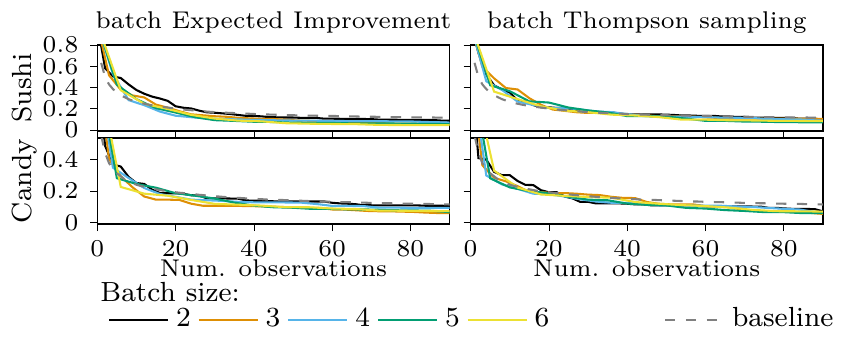}
\caption{Summary of the performance of the presented method on the real datasets. Each plot illustrates the minimum value seen so far as a function of the number of function evaluations. All data sets  are scaled between 0 and 1. Different colors in plots are different batch sizes, rows are different data sets and columns are different acquisition functions. Each line is a mean of 10 different runs. The dashed black line shows the average performance of the baseline, random search. All lines use \ep as an inference method.}
\label{fig:summary}
\end{figure}

Figure \ref{fig:summary} shows the minimum function value seen so far as a function of the number of function evaluations. The results show batch sizes 2--6 and three acquisition functions for all 4 datasets when \ep is used for inference. Each line shows the average over 10 random runs for each setting. The results are consistent with the results of the synthetic functions from the Sigopt library. One notable difference compared to the simulated results is that all methods beat the baseline only barely due to the low signal-to-noise ratio. When fitting a \gp to the data sets that are scaled between 0--1, the noise standard deviation varies between 0.1--0.2. The continuous features alone seem not to be enough for performing the optimization.

\section{CONCLUSION}
\label{sec:conclusion}

The results suggest that batch winner feedback is sufficient for optimization tasks as it is easier to collect than the full ranking of the batch and it performs almost as well in \bo. The results suggest that if the dimensionality of the data is low and the chosen batch size is small, practitioners should use \ep inference or \mcmc sampling to approximate the posterior distribution.
As an acquisition function, we recommend \qei for low dimensions and small batch sizes and \ts for high dimensions or large batch sizes due to its scalability.

Noise poses a problem to the presented approaches. Preferential observations are inherently less informative than direct observations. Noise reduces the information carried by preferential queries even more. Larger batch sizes do not reduce the problem due to the need to jointly account for the uncertainty of the whole batch. We see no apparent solution to this problem and it, unfortunately, reduces the usability of preferential feedback in noisy problems. With highly noisy data, random search with large batches might be a good option.

\section{Acknowledgements}
We acknowledge the computational resources provided by the Aalto
Science-IT project and support by the Academy of Finland Flagship
programme: Finnish Center for Artificial Intelligence, FCAI.

\label{sec:ref}
\bibliographystyle{IEEEbib}
\bibliography{pbbo}

\onecolumn

\renewcommand{\thesubsection}{\Alph{section}\arabic{subsection} }

\renewcommand{\thesection}{\Alph{section}}

\setcounter{equation}{7}
\setcounter{figure}{5} 
\setcounter{section}{0}

\section{Deriving the likelihood of the batch winner}
If the provided batch type feedback is the batch winner $\x_j$ ,
\begin{equation*}
\mathbf{C} = \left[ \begin{matrix}
j & 1 \\
\vdots & \vdots \\
j & j-1 \\
j & j+1 \\
\vdots & \vdots \\
j & q \\
\end{matrix} \right]
\end{equation*}

the likelihood can be simplified as  
\begin{eqnarray}
 p(\C | \f) = &  \int \ldots \int \left( \prod_{i=1, i \neq j}^m \mathbbm{1}_{y_{j}  \leq  y_i} \right) \left( \prod_{k=1}^q N(y_k| f_k, \sigma^2) \right)  \diff y_1  \ldots \diff y_q\\ =&\int N(y_j| f_j, \sigma^2) \prod_{i=1, i \neq j}^q \left( \int_{y_j}^\infty N(y_i| f_i, \sigma^2) \id y_i  \right) \id y_j \\ =& \int N(y_j| f_j, \sigma^2) \prod_{i=1, i \neq j}^q \Phi \left( \frac{y_j-f_i}{\sigma} \right)  \diff y_j.
\end{eqnarray}

\section{Details of the expectation propagation model}
In expectation propagation, the observation model is approximated by a product of distributions from the exponential family,
\begin{equation}
p( \{ \mathbf{C}^b \}_{b=1}^B \given \{ \f^b \}_{b=1}^B)  = \prod_{b=1}^B p( \mathbf{C}^b \given \f^b ) \approx \prod_{b=1}^B q_b(\f^b)  = q(\{ \f^b \}_{b=1}^B).
\end{equation}
The approximative distributions, $q_b(\cdot)$, (parametrized by distributions from the exponential family) are called 'site approximations'. The prior distribution and the product of the site approximations can be used to approximate the posterior of the latent function values,
\begin{equation}
q(\f) = \frac{\Normal{\f \given \mathbf{0}, k_{\boldsymbol{\theta}}(\X,\X)} \left( \prod_{b=1}^{B} q_b(\f^b)\right) }{\int \cdots \int  \Normal{\f \given \mathbf{0}, k_{\boldsymbol{\theta}}(\X,\X)} \left( \prod_{i=b}^{B} q_b(\f^b)\right) \id \f^1 \cdots  \id \f^{B}}, 
\end{equation}
where $k_{\boldsymbol{\theta}}(\cdot, \cdot)$ is the covariance function parametrized by hyper-parameters $\theta$, and $\X = [\X^1,\ldots,\X^B]\tr$. Before going into how to optimize the parameters of the site approximations, let us first define the 'cavity distribution',
\begin{equation}
q_{-j}(\f^j) \propto \int \cdots \int  \Normal{\f \given \mathbf{0}, k_{\boldsymbol{\theta}}(\X,\X)} \left( \prod_{\substack{b=1 \\ b\neq j}}^{B} q_b(\f^b)\right) \id \f^1 \cdots \id \f^{j-1} \id \f^{j+1} \cdots \id \f^{B}.
\end{equation}
Also, let us define the 'tilted distribution', which is the cavity distribution multiplied with the exact likelihood of the missing observation
\begin{equation}
q_{\backslash j}(\{ \f^j \}_{b=1}^B) = p( \mathbf{C}^j \given \f^j ) q_{-j}(\f^j).
\end{equation}
In expectation propagation, we iteratively try to optimize the site approximations as a part of the global approximation. This is done by minimizing the following Kullback-Leibler (KL) -divergence 
\begin{equation}
\text{KL}\left( q_{\backslash j}(\f^j ) \,||\, q_{-j}(\f^j  q_j (f^j) \right) = \int q_{\backslash j}( \f^j ) \log \left( \frac{q_{\backslash j}(\f^j )}{ q_{-j}(\f^j)  q_j (f^j)} \right) \diff \f^j,
\end{equation}
with respect to the parameters of the j:th approximative distribution. The solution to this equals to matching the zeroth, first and second moments of the global approximation and the tilted distribution\footnote{Seeger, M. Expectation propagation for exponential families. Technical report, 2005.}. Since in our case the moments of the tilted distribution are not computable in closed form, we approximate them by sampling.

\section{Details of the variational inference model}
The idea in variational inference is to minimize the Kullbak-Leibler -divergence between the approximative distribution and the intractable posterior,
\begin{equation}
\text{KL}\left( q(\{f^b\}_{b=1}^B) \,||\, p(\{f^b\}_{b=1}^B \given \{ \X^b \}_{b=1}^B, \{\mathbf{C}^b\}_{b=1}^B ) \right),
\end{equation} 
with respect to the parameters of the approximative distribution. In our case, the approximative distribution is defined either by Equation (7) or (8). However, since the posterior is intractable, the above equation must be approximated my maximizing the evidence lower bound,
\begin{equation}
\E_{q(\{f^b\}_{b=1}^B)}\left[ \log p( \{\mathbf{C}^b\}_{b=1}^B \given \{f^b\}_{b=1}^B ) \right] - \text{KL}\left( q(\{f^b\}_{b=1}^B) \,||\, p(\{f^b\}_{b=1}^B \given \{ \X^b \}_{b=1}^B )  \right).
\end{equation}
This requires some gradient based optimization scheme. The required derivatives can be found e.g. from Titsias (2016)\footnote{Titsias, M. K. One-vs-each approximation to softmax for scalable estimation of probabilities. In Proceedings of the 29th Conference on Neural Information Processing Systems, pp. 4161--4169, 2016.}.

\section{Explaining the high computational cost of pq-EI}
As we further open Equation (8), it becomes:
\begin{align*}
\text{pq-EI} =& \E_{\y,y_{min} } \left[ \left( \max _{i \in [1, \ldots , q]} \left(y_{min} - y_i \right) \right)_+ \right] \\
=& \int_{-\infty}^{\infty} p(y_{min}) \sum_{i=1}^q \E_{\y} \left( y_{min} - y_i \middle | y_i \leq y_{min},\; y_i \leq y_j \; \forall j \neq i,\;y_{min}  \right) \times  p(y_i \leq y_{min},\; y_i \leq y_j \forall j \neq i \;|\; y_{min} ) \id y_{min},
\end{align*}
where $p(y_{min})$ is the distribution of minimum of $p\times q$ dimensional Normal distribution\footnote{Arellano-Valle, R. B., and Genton, M. G. On the exact distribution of the maximum of absolutely continuous dependent random variables. Statistics \& Probability Letters, 78(1), 27-35, 2008.} ($p$ is number of iterations before this and q is the batch size). More explicitly
\begin{equation*}
p(y_{min}) = \sum_{i=1}^p \sum_{j=1}^{q} \Normal{-y_{min}\;|\; -\mu_{ij}, \Sigma_{ij\;ij}} \boldsymbol{\Phi}_{pq-1}\left(-y_{min} \mathbf{
1} \;| \; -\boldsymbol{\mu}_{-ij}(y_{min}),\; \Sigma_{-ij-ij,ij}\right),
\end{equation*}
where $\mu_ij$ is the posterior mean of the latent function at $i$:th batch and $j$:th batch location, $\Sigma_{ij\;ij}$ is the posterior covariance of predictive output at the same location and 
\begin{equation*}
\boldsymbol{\mu}_{-ij} (y_{min}) = \boldsymbol{\mu}_{-ij} - ( y_{min} - \mu_{ij})  \boldsymbol{\Sigma}_{-ij\;ij} / \Sigma_{ij \; ij},\text{ and } \Sigma_{-ij\;-ij,ij} = \Sigma_{-ij\; -ij} - \boldsymbol{\Sigma}_{-ij \; ij} \boldsymbol{\Sigma}_{-ij \; ij}\tr / \Sigma_{ij\; ij}.
\end{equation*}
Furthermore $\E_{\y} \left( y_{min} - y_i \middle | y_i \leq y_{min},\; y_i \leq y_j \; \forall j \neq i,\;y_{min}  \right)$ can be computed efficiently with Tallis formula for small batch sizes. However, even thought we need to only perform numerical integration over one dimension (as multidimensional cumulative normal distributions can efficiently be approximated), the computation of $p(y_{min})$ becomes computationally very demanding because of the need for computation of very high dimensional cumulative normal distribution functions ($pq-1$ becomes very large after few iterations).

\section{Details and additional results for Section 4.1}
\begin{figure}[tb!]
	\centering
    \includegraphics{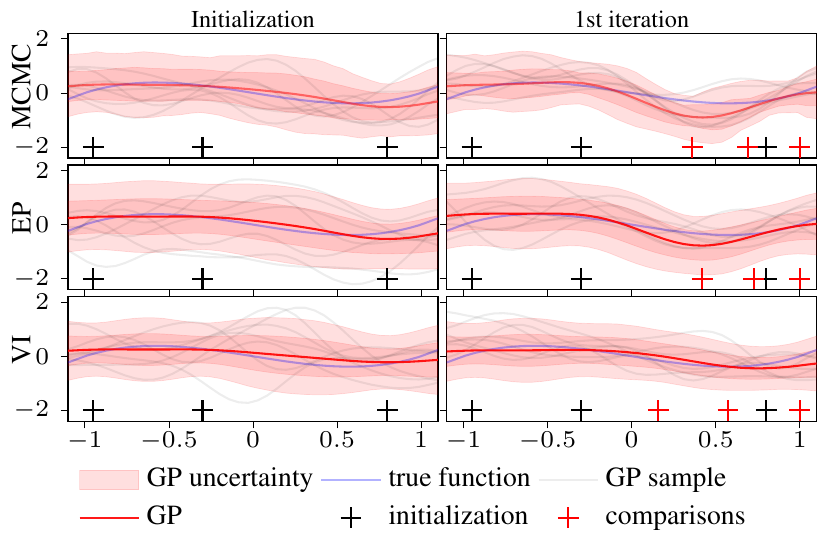}
\caption{Same as in Figure 1, but for batch size 3.  \label{fig:06} }
\end{figure}

\begin{figure}[tb!]
	\centering
    \includegraphics{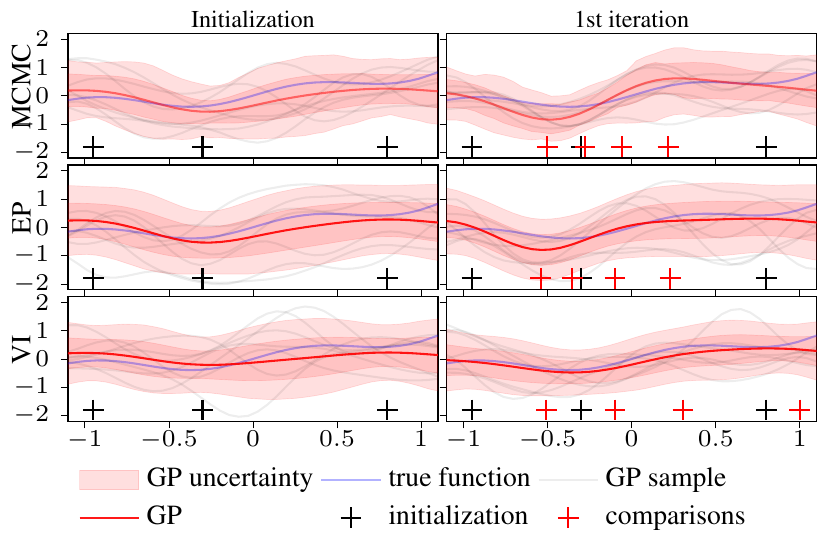}
\caption{Same as in Figure 1, but for different function. \label{fig:33} }
\end{figure}

\begin{figure}[tb!]
	\centering
    \includegraphics{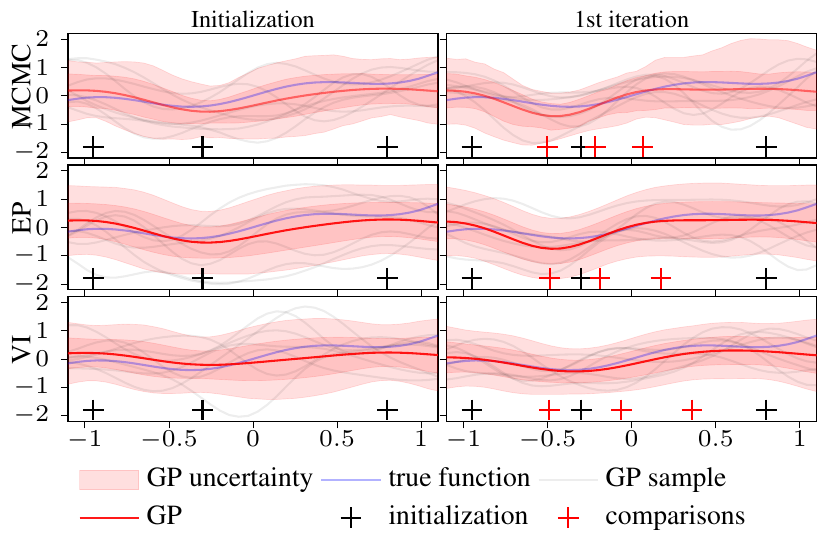}
\caption{Same as in Figure 1, but for batch size 3 and for different function. \label{fig:36}}
\end{figure}

Figure \ref{fig:06} presents same experiment as in Section 4.1, but for batch size of 6.  Figures \ref{fig:33} and \ref{fig:36} have results for function $\frac{1}{4}\sin(5x)+\frac{1}{2}e^x-\frac{1}{2}$.

\clearpage
\section{Details and additional results for Section 4.2}
For all simulation runs, the function bounds and output was scaled between 0 and 1, for all dimensions. The batch feedbacks were computed from outputs which were corrupted with noise that has standard deviation of $0.05$. \qei was computed with 5000 posterior samples. All acquisition functions were optimized using limited memory  Broyden–Fletcher–Goldfarb–Shanno (BFGS) algorithm with box constraints. The optimization was restarted 30 times for \qei. To increase the robustness of the \ep, we do not allow distances between points to be less than 0.05 within a single batch. The numerical gradients of \ts were computed with $\delta=10^{-5}$ and only 100 closest samples were taken into account when conditioning on the evaluated samples. Optimization of \ep was limited to at maximum 100 iterations. Optimization of \mfvi was limited to 50 iterations and Adam was used for optimization.

\begin{figure}[tb!]
    \centering
    \includegraphics{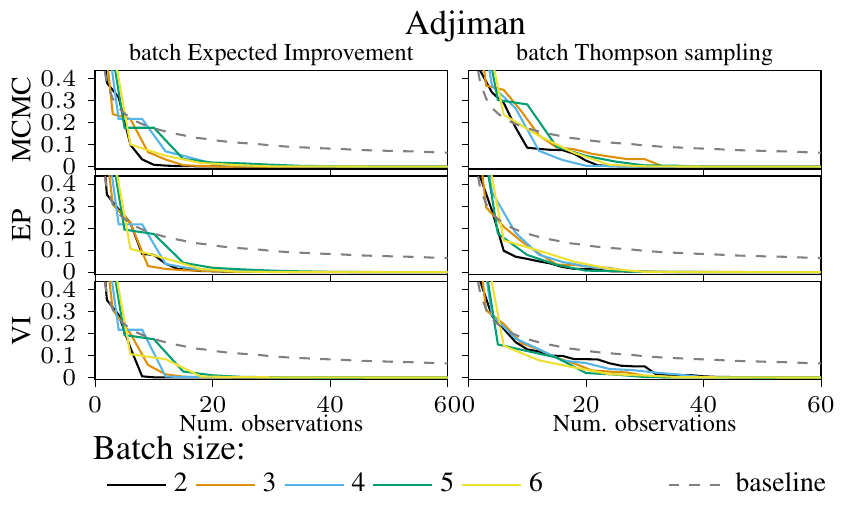} 
    ~
    \includegraphics{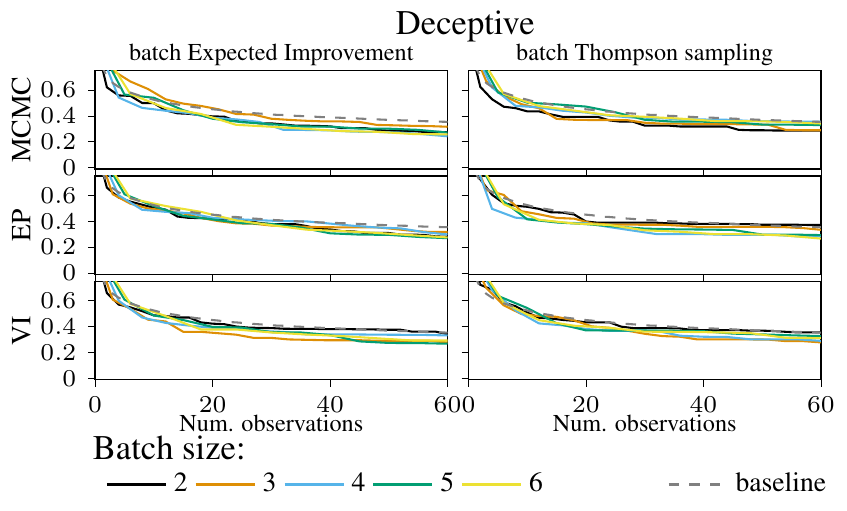}
    \\
    \includegraphics{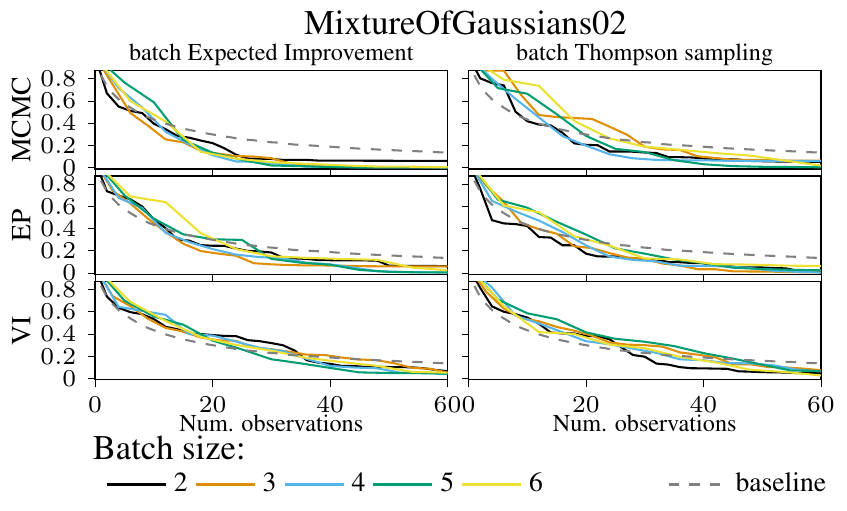}
    ~
    \includegraphics{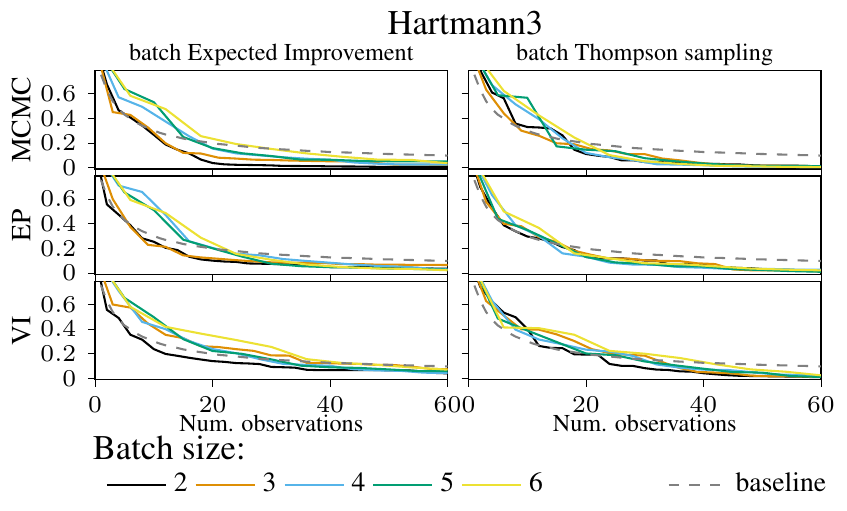}
    \\ 
    \includegraphics{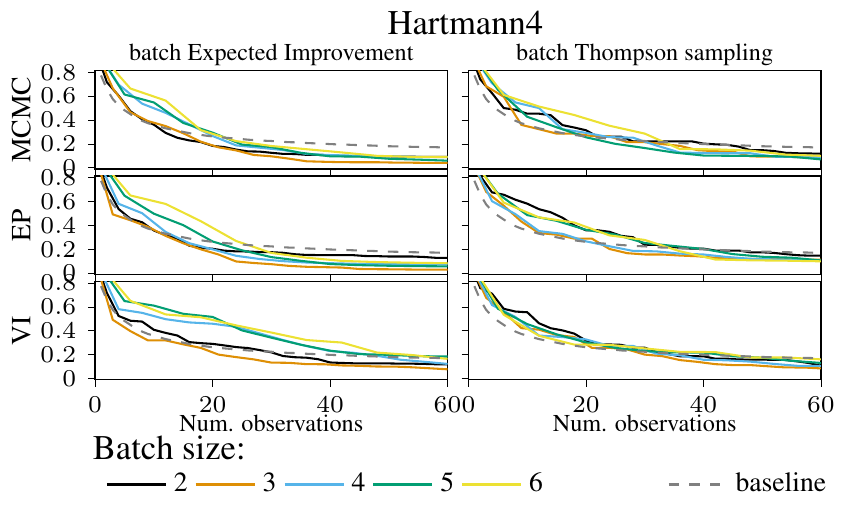}
    ~ 

\caption{Same as in Figure 2, but for Adjiman, Deceptive, MixtureOfGaussians02 and 3 and 4 dimensional Hartmann-functions from the Sigopt function library.}
\label{fig:sigopt}
\end{figure}

Figure \ref{fig:sigopt} presents similar results as in Section 4.2 for the 5 other functions from the Sigopt function library. All these functions are well known global optimization bench mark functions.

\clearpage
\section{Additional results for Section 4.3}
\begin{figure}[tb!]
    \centering
    \includegraphics{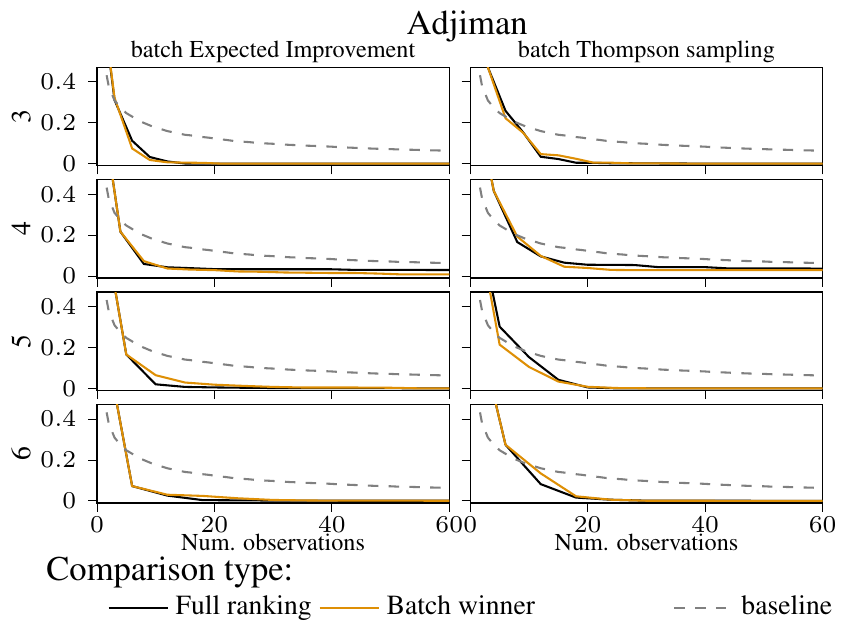} 
    ~
    \includegraphics{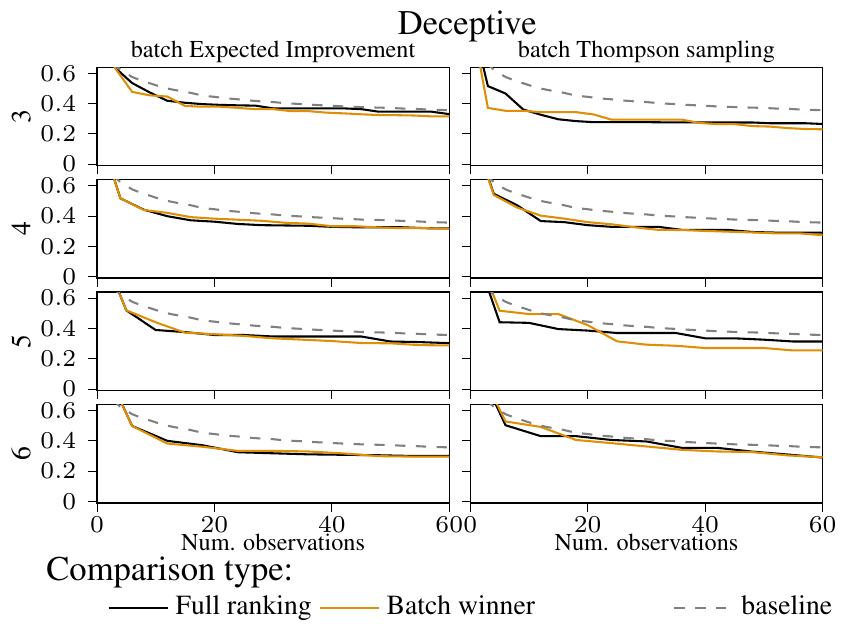}
    \\
    \includegraphics{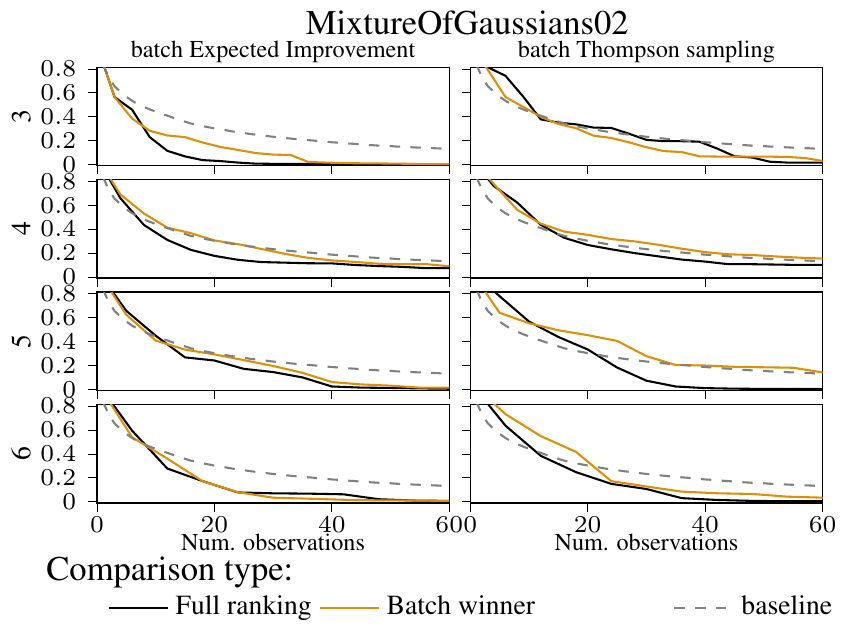}
    ~
    \includegraphics{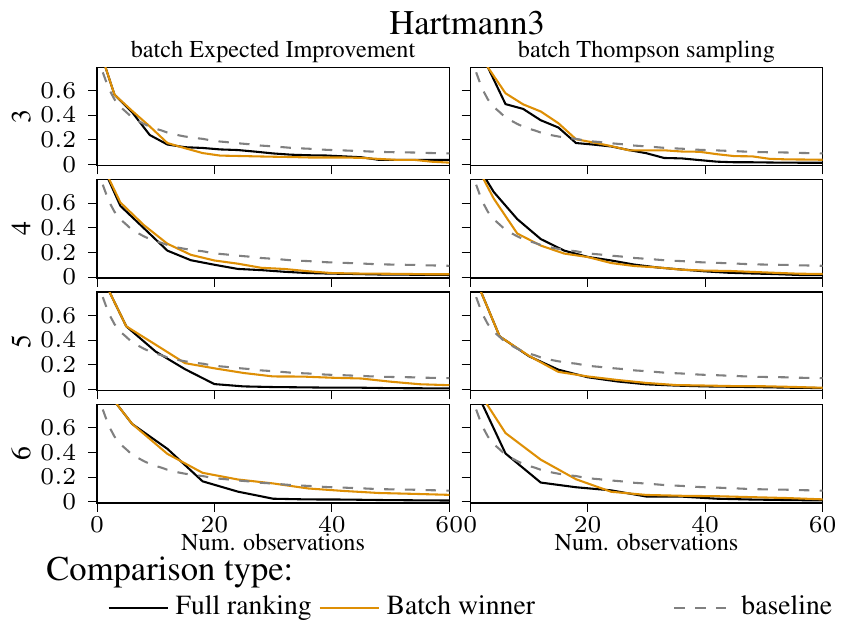}
    \\ 
    \includegraphics{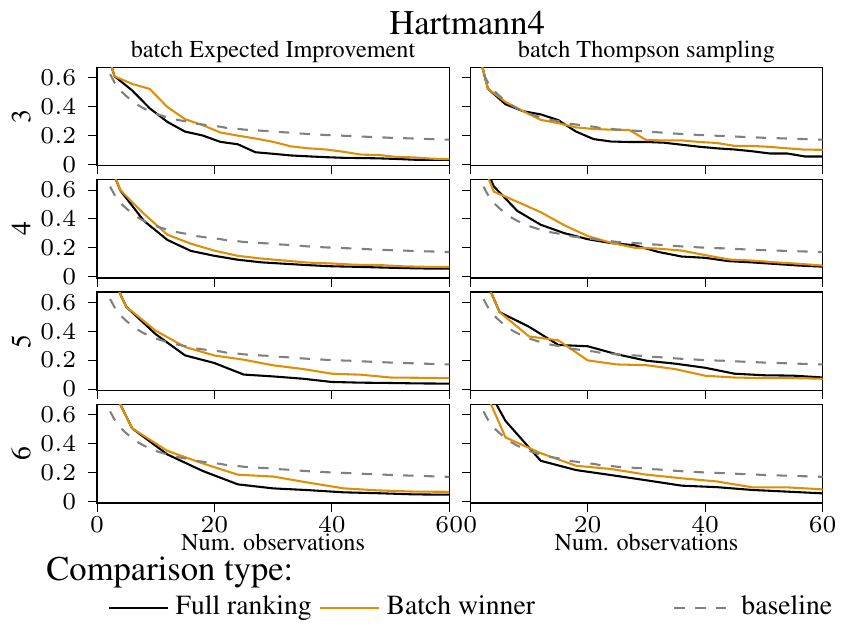}
    ~ 
    \includegraphics{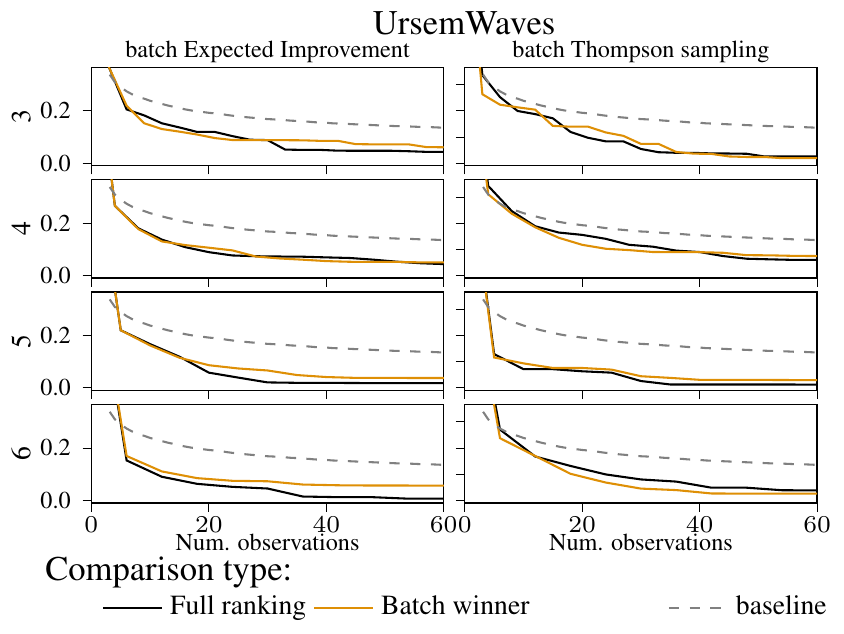}
    
\caption{Comparison of full ranking and batch winner feedbacks for six functions from the Sigopt function for the tree acquisition functions and batch sizes 3--6.}
\label{fig:sigopt_mcmc}
\end{figure}

Additional results for the experiments in Section 4.3. Figure \ref{fig:sigopt_mcmc} compares full ranking and batch winner feedbacks for six functions from the Sigopt library for \qei and \ts for batch sizes 3--6.

\clearpage
\section{Additional results for Section 4.4}
The details of the experiments are the same as for the experiments in Section 4.4 with few exceptions. Since the functions are higher dimensional, acquisition optimization for \qei was restarted 60 times. Also, no noise was added to real data.

\begin{figure}[bt!]
    \centering
    \includegraphics{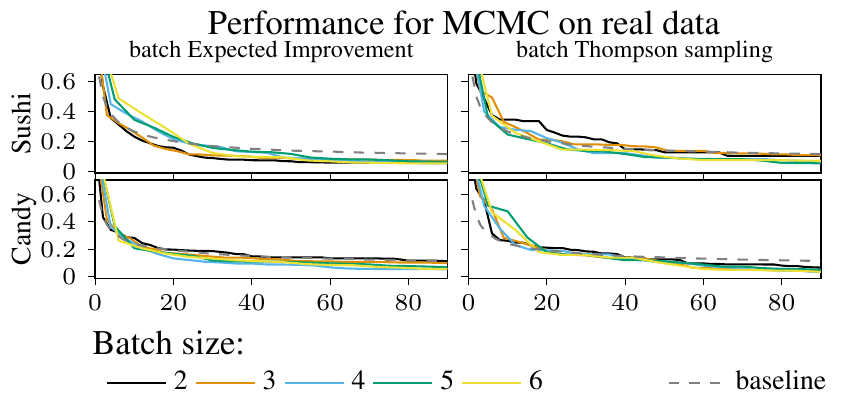} 
    ~
    \centering
    \includegraphics{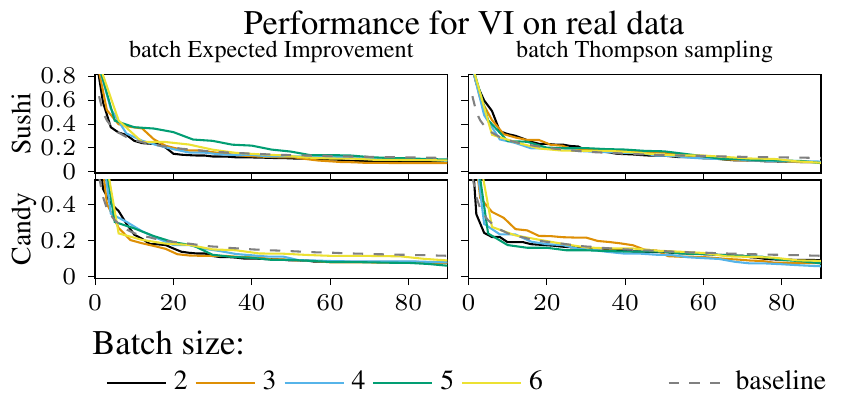}
    \\
\caption{Performances of \mcmc and \mfvi inferences on the ral data.}
\label{fig:real_inf}
\end{figure}

\begin{figure}[tb!]
	\centering
    \includegraphics{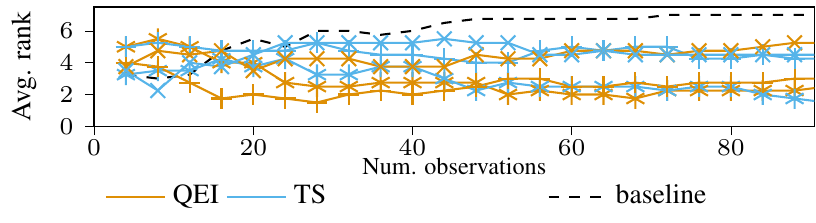}
    \noindent\\
    \vspace{-1.9mm}
    \hbox{\hspace{57mm}\includegraphics{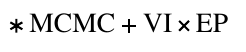}}
\caption{Same as in Figure 3, but for Sushi and Candy datasets.}
\label{fig:real}
\end{figure}

Figure \ref{fig:real_inf} presents similar results as in Section 4.4 for \mfvi and \mcmc. Figure \ref{fig:real} illustrates the ranks of different inference methods similarly as in Figure 3, but for Sushi and Candy datasets.

\end{document}